%% file: ACII2023-template.tex
\documentclass[conference]{IEEEtran}
\IEEEoverridecommandlockouts
\usepackage{cite}
\usepackage{amsmath,amssymb,amsfonts}
\usepackage{algorithmic}
\usepackage{graphicx}
\usepackage{textcomp}
\usepackage{xcolor}
\usepackage{booktabs}
\usepackage{subfigure}
\usepackage{multirow}
\usepackage{authblk}
\usepackage[symbol]{footmisc}

\def\BibTeX{{\rm B\kern-.05em{\sc i\kern-.025em b}\kern-.08em
    T\kern-.1667em\lower.7ex\hbox{E}\kern-.125emX}}
\begin{document}

\title{Context Unlocks Emotions: Text-based Emotion Classification Dataset Auditing with Large Language Models}


\makeatletter
\newcommand\email[2][]%
   {\newaffiltrue\let\AB@blk@and\AB@pand
      \if\relax#1\relax\def\AB@note{\AB@thenote}\else\def\AB@note{\relax}%
        \setcounter{Maxaffil}{0}\fi
      \begingroup
        \let\protect\@unexpandable@protect
        \def\thanks{\protect\thanks}\def\footnote{\protect\footnote}%
        \@temptokena=\expandafter{\AB@authors}%
        {\def\\{\protect\\\protect\Affilfont}\xdef\AB@temp{#2}}%
         \xdef\AB@authors{\the\@temptokena\AB@las\AB@au@str
         \protect\\[\affilsep]\protect\Affilfont\AB@temp}%
         \gdef\AB@las{}\gdef\AB@au@str{}%
        {\def\\{, \ignorespaces}\xdef\AB@temp{#2}}%
        \@temptokena=\expandafter{\AB@affillist}%
        \xdef\AB@affillist{\the\@temptokena \AB@affilsep
          \AB@affilnote{}\protect\Affilfont\AB@temp}%
      \endgroup
       \let\AB@affilsep\AB@affilsepx
}
\makeatother
\author[1*]{Daniel Yang}
\author[1*]{Aditya Kommineni}
\author[1,2]{Mohammad Alshehri}
\author[1]{Nilamadhab Mohanty} 
\author[1]{\\ Vedant Modi}
\author[1]{Jonathan Gratch}
\author[1]{Shrikanth Narayanan}

\affil[ ]{
\textit {\{dyang165,akommine\}@usc.edu},
\textit{mohammed.shehri.32@aramco.com},
}

\affil[ ]{
\textit {\{nmohanty,vkmodi\}@usc.edu},
\textit{gratch@ict.usc.edu},
\textit{shri@ee.usc.edu}
}

\affil[1]{University of Southern California}
\affil[2]{Saudi Aramco}
\maketitle

\begin{abstract}
\input{Abstract}
\end{abstract}

\footnotetext[1]{equal contribution.}
\section{Introduction}
\input{Introduction}

\section{Related Work\label{section2}}
\input{relatedwork}

\section{Prompting LLMs for Context Generation}
\input{context}

\section{Datasets}
\input{dataset}

\section{Experiments}
\input{experiments}
\subsection{Objective Evaluation}
\input{objectiveEval}

\subsection{Subjective Evaluation}
\input{subjectiveEval}

\section{Results}
\input{Tables/results_bert_sent}
\input{results}

\section{Conclusion}
\input{conclusion}

\section*{Ethical Impact Statement}
\input{ethicalImpact}

\section{Acknowlegements}
This material is based upon work supported by the National Science Foundation Graduate Research Fellowship under Grant No. DGE-1842487. We would like to James Hale for providing us with Amazon MTurk account for completion of our evaluation.

\bibliographystyle{IEEEtran}
\bibliography{ref}



\end{document}

%% file: Abstract.tex
The lack of contextual information in text data can make the annotation process of text-based emotion classification datasets challenging. As a result, such datasets often contain labels that fail to consider all the relevant emotions in the vocabulary. This misalignment between text inputs and labels can degrade the performance of machine learning models trained on top of them. As re-annotating entire datasets is a costly and time-consuming task that cannot be done at scale, we propose to use the expressive capabilities of large language models to synthesize additional context for input text to increase its alignment with the annotated emotional labels. In this work, we propose a formal definition of textual context to motivate a prompting strategy to enhance such contextual information. We provide both human and empirical evaluation to demonstrate the efficacy of the enhanced context. Our method improves alignment between inputs and their human-annotated labels from both an empirical and human-evaluated standpoint.

%% file: Introduction.tex
Text-based emotion classification is the task of assigning relevant emotion categories to input text samples. A sample can be given either a single label \cite{isear,dd,tweeteval,semeval2019t3} or multiple labels \cite{goemo}. For instance, the SemEval 2019 Task 3 dataset\cite{semeval2019t3} contains textual dialogues labeled with four categorical labels: happy, sad, angry, or others. On the other hand, the GoEmotions dataset features a broader range of labels, including 27 discrete emotion categories and one neutral label. Due to the differing number of emotion categories and the adopted annotation protocol, the same emotion label may have different significance across datasets. For example, if a dataset has a label for anger but not digust, annotators may associate the feeling of disgust with the label of anger. However, if a dataset has both anger and disgust, these feelings would be mapped to the appropriate construct. This makes curating consistent text-based emotion classification datasets rather challenging.

\input{Tables/ambiguous_exmples}

Despite the efforts to create high-quality text emotion classification datasets\cite{dd,goemo}, they still face problems in the annotation process. In particular, texts from uni-label datasets are constrained to a single label, when they could actually reflect multiple emotions. Similarly, multi-label datasets may not cover all emotion labels relevant to the text input. We notice that this problem is exacerbated when textual inputs contain only a few words. Table~\ref{tab:examples} includes sentences from the GoEmotions\cite{goemo} dataset demonstrating such labeling inaccuracies. For each text input, we show the provided label, along with other possible emotions that the text may reflect. For example, the text ``Wow!!!" is labelled with excitement and surprise, but can reasonably be associated with amusement, joy, admiration, anger, annoyance, curiosity, nervousness, optimism, or realization. Re-annotating datasets to resolve these issues is time-consuming as well as costly in terms of resources. Furthermore, without additional context, such problematic data frequently delivers ambiguous information, making it challenging to define a universally-agreed-upon labeling standard. As a solution, we suggest using Large Language Models (LLMs) such as GPT-3.5 and GPT-4\cite{gpt4} in a scalable and automatic method to audit pre-existing text-based emotion classification datasets.

Large Language Models (LLMs), which are based on the Transformer~\cite{10.5555/3295222.3295349} architecture, have been shown to excel in various natural language processing (NLP) tasks \cite{gpt3}. Transformers are sequence-to-sequence models that process discrete tokens. For NLP tasks, they are often pre-trained through a self-supervised denoising objective, where a token is masked in the input sequence and then predicted by the model. GPT-3.5 and GPT-4 are specific LLMs designed to predict the next word in a sequence. When predicting a token, the model masks the tokens to its right, focusing only on the preceding tokens. This process allows the model to generate original text by sampling the next token based on previously generated tokens, in an autoregressive manner. Furthermore, GPT-3.5 and GPT-4 are trained with reinforcement learning from human feedback \cite{instructgpt}, which employs human rankings of model outputs to better align text generation with how humans would respond.

The primary method to evaluate a moderately-sized language model (e.g., BERT~\cite{bert}) on a downstream task consists of a two-step process: pretraining and fine-tuning. Initially, the model undergoes self-supervised pretraining on a large dataset, followed by fine-tuning on a smaller, target dataset. However, due to the enormous number of parameters in GPT-3.5 and GPT-4, fine-tuning these models is often impractical. A recent alternative for applying these models to downstream tasks is prompting, i.e., designing input words, tokens, or embeddings such that the outputs can be applied directly to downstream tasks without fine-tuning. Typically, the generated words are mapped to labels which are then used to perform zero-shot or few-shot evaluation.

Previous work has prompted LLMs for text-based emotion classification \cite{jackofalltradeschatgpt}, but their performance significantly lags behind a fine-tuned language model \cite{goemo, emotionx}. However, both GPT-3.5 and GPT-4 have been trained on a large variety of text from various online sources, including numerous emotionally-rich texts. To accurately predict the next tokens in emotionally-rich texts, these models need to develop an understanding of semantics of human emotions \cite{memobert}. In this paper, we propose a strategy to combine the benefits of pre-training/fine-tuning, and prompting. In particular, we use GPT-3.5 and GPT-4 prompting as a method to audit pre-existing text-based emotion classification datasets by evaluating existing context and generating new context when it is inadequate. Ultimately, we improve the alignment between input texts and labels of the dataset, so that it is easier to fine-tune language models and learn the relationship between them. This paper has three main contributions, summarized below: 
\begin{itemize}
    \item We propose a formal definition of textual context to motivate a prompting strategy. The prompting strategy is designed to generate appropriate context, as specified by the definition.
    \item We present results from human-provided feedback on surveys to show that our inputs and labels are well-aligned from a human perspective.
    \item We demonstrate empirical improvements in classification performance on text-based emotion classification with our modified datasets compared to using the original GoEmotions dataset. 
\end{itemize}

%% file: Tables/ambiguous_exmples.tex
\begin{table}[]
\caption{Ambiguous examples in GoEmotions dataset}
\label{tab:examples}
\begin{tabular}{ccc}
\toprule
\textbf{Text Input} & \textbf{Annotated Labels}      & \textbf{Other Possible Labels} \\ \midrule
Noooo not the booze                                                      & neutral              & \begin{tabular}[c]{@{}c@{}}sadness, annoyance,\\ anger disappointment\end{tabular}                                                         \\ \hline
Calm down bro                                                            & neutral              & \begin{tabular}[c]{@{}c@{}}annoyance, caring,\\ nervousness, anger\end{tabular}                                                                   \\ \hline
\begin{tabular}[c]{@{}c@{}}I’m not crying, \\ you’re crying\end{tabular} & annoyance            & \begin{tabular}[c]{@{}c@{}}sadness, grief\\ embarrassment\end{tabular}                                                                     \\ \hline
Wow!!!                                                                   & excitement, surprise & \begin{tabular}[c]{@{}c@{}}amusement, joy\\ admiration, anger\\ annoyance, curiosity, \\ nervousness, optimism,\\ realization\end{tabular} \\ \bottomrule
\end{tabular}
\end{table}

%% file: relatedwork.tex

\subsection{Text-based Emotion Classification}
Text-based emotion classification is a well-established task with multiple existing datasets \cite{isear,dd,tweeteval,semeval2019t3}. However, some remaining challenges in the domain, as expressed in \cite{TER_Challenges}, include fuzzy emotional boundaries, incomplete extractable emotional information in texts, and lack of large scale datasets. The primary goal of this study is to assess the degree of contextual information available within input sentences, and subsequently enhance the contextually deficient sentences using text prompt techniques, leveraging large language models as a resource.

With the release of Large Language Models (LLMs) such as GPT-3.5 and GPT-4, there has been growing interest in assessing their Natural Language Understanding and Generation ability. \cite{jackofalltradeschatgpt} provides a comprehensive evaluation of ChatGPT, demonstrating its semantic understanding of text and competitive zero-shot performance to fully supervised models on diverse NLP tasks. Additionally, \cite{gilardi2023chatgpt} highlights the efficiency of LLMs at performing manual data annotations on NLP tasks, both in terms of incurred cost and efficacy. Particularly, this work demonstrates that ChatGPT outperforms crowd-sourced workers  in topic detection, stance detection and policy frame detection on tweet data.

However, to the best of our knowledge, this work constitutes the first work improving  the disambiguity between the emotional content of the sentence and the corresponding emotion label based on available contextual information.

\subsection{Context-based Affect Recognition}
Context has a significant impact on how we understand and evaluate human emotions across modalities such as images, text and video. Facial expressions have shown to be unreliable predictors of experienced emotions in the absence of contextual information  \cite{barrett2019emotional}. \cite{mittal2020emoticon}, \cite{lee2019context} and \cite{kosti2017emotion} propose methods to incorporate visual context in image and video modalities to improve emotion recognition. In NLP, most works have investigated context for emotion detection in text-based conversations. In this case, context is defined as the utterances in the prior turns, and by secondary parties in the conversation; for example, \cite{liscombe2005using}, \cite{Lee2005Towarddetectingemotionsin} and \cite{zhang2019modeling} have proposed approaches to embed context in conversational settings into the emotion detection process. We believe that this definition of context needs to be further qualified for our use case. 

\subsection{Auditing Datasets}
Several forms of dataset auditing have been explored by the artificial intelligence community. A widely used method is data augmentation, where multiple transformations are applied to the inputs to increase the number of plausible views of the input data. However, our approach fundamentally differs from this method because data augmentation does not aim to fix poor alignment between the inputs and the output labels. To the best of our knowledge, two recent works attempted to do this. The first is confident learning \cite{confident}, which models the label noise in image datasets to predict label errors. The second is related to the LAION-5B dataset curation \cite{laion5b}, which uses CLIP~\cite{clip} as a similarity score to filter misaligned (image, text) pairs. Our approach builds on top of these ideas by automatically modifying the input to fit the labels better when there is misalignment.

%% file: context.tex
\begin{figure*}
    \centering
    \includegraphics[width=\linewidth]{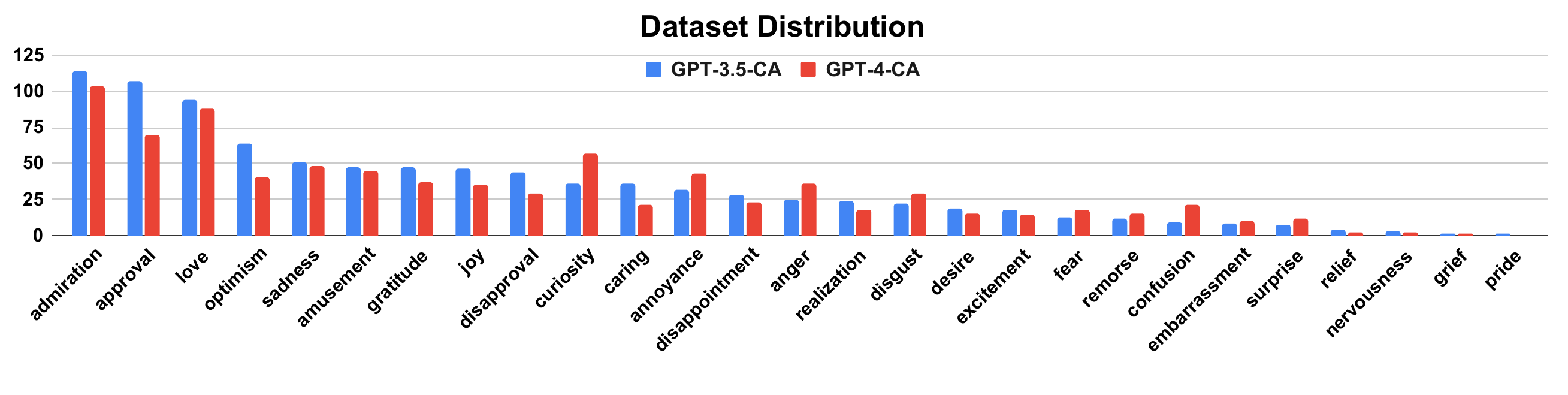}
    \vspace{-1cm}
    \caption{Distribution of emotions in GPT-3.5-CA and GPT-4-CA. Additionally, GPT-3.5-CA and GPT-4-CA have 357 neutral and 407 neutral inputs, respectively}
    \label{fig:emo_dist}
\end{figure*}
In order to create an effective strategy for generating contextual prompts, our first task is to define textual context properly. This is essential as it provides guidelines to motivate the prompting scheme. In this section, we discuss the approach used for prompting GPT-3.5 and GPT-4, as well as the efficacy of the prompts with respect to our definition of textual context.

\subsection{What is appropriate textual context for emotion classification?}
We describe appropriate textual context as words added to an input such that the modified input conveys emotions which are \textit{faithful} and \textit{unambiguous} for a target audience. The key idea in our definition is that the value of context-conveying text depends on human interpretation. When people read a piece of text, they associate it with their own background knowledge, experiences, and emotions, which leads them to make interpretations under a personal lens. These interpretations can be diverse or similar, depending on the amount of context the input text already contains. For instance, consider the example ``Noooo not the booze" from Table~\ref{tab:examples}, which is sourced from the training split of the GoEmotions dataset. Some interpretations may include ``You are at a party, but the booze spilled," ``You finished the booze, but you want more," and ``Your friend asked if you want booze or soda." For a given target audience, let us denote the set of these interpretations as $I$. $I$ may depend on demographic factors such as age or gender. Moreover, $I$ does not depend on the author's original intent, but rather on how the target audience perceives the text. Each interpretation in $I$ can be mapped to emotion labels, the specifics of which change depending on the underlying representation of emotion that is used (e.g., discrete labels, valence/arousal). For the interpretations ``you are at a party, but the booze spilled" and ``You finished the booze, but you want more," the GoEmotions label that best fits, in our opinion, is sadness, disappointment and disapproval. For the interpretation "Your friend asked if you want booze or soda," the best GoEmotions label is neutral.

For the context to be appropriate, it must achieve two objectives. First, the context-added text must be \textit{faithful}. We say that a context-added text is faithful if its set of interpretations is a subset of the set of interpretations of the original text. For example, consider the following context-added text, generated from GPT-4: ``Noooo, not the booze. It's just that I would prefer something non-alcoholic to drink". The set $I'$ of interpretations for this text includes variations of ``Someone asked whether you want booze or soda". Since $I'$ is a subset of $I$, the context is faithful. However, consider the silly GPT-4 generated context of ``Noooo, not the booze. I can't handle it, I'm just a little teapot!" Then, an interpretation in $I'$ is ``You are a teapot that does not want to hold alcohol". It is safe to say that this is not a part of $I$, so the context is not faithful. This condition prevents the context from being a wall of text that overwrites the significance of the original text from the perspective of the target audience.

On top of being faithful, a context-added text must be \textit{unambiguous}. For the given text to be unambiguous, the assigned emotion labels must match across all interpretations. This means that two interpretations should not have differing emotion labels. In the case of the original text, the set $I$ has elements which result in two GoEmotions labels so it is not unambiguous. On the other hand, the interpretations for the text ``Noooo, not the booze. It's just that I would prefer something non-alcoholic to drink" all are universally associated with the neutral emotion. Therefore, we can conclude that this context-enhanced text is unambiguous.

A limitation of our approach is that it is not possible to test for these metrics empirically. Instead, we use our definition to motivate our prompting scheme so that the context-added text is more likely to be faithful and unambiguous.

\subsection{Prompting For Context Generation and Context Evaluation}
Our objective is to design prompts that allow LLMs to generate contextual messages from the input sentences, ensuring the context-enhanced text conveys the intended emotions accurately and unambiguously while keeping the original message faithful. We achieve this by wording our prompts carefully. The prompt we used is as follows:

    \textbf{Prompt:} ``system": You are a Reddit user editing your post. You are banned from using these words, or any forms of them: admiration, amusement, anger, annoyance, approval, caring, confusion, curiosity, desire, disappointment, disapproval, disgust, embarrassment, excitement, fear, gratitude, grief, joy, love, nervousness, optimism, pride, realization, relief, remorse, sadness, surprise, ``user": Add one or two sentences to this Reddit post to convey the emotions of \_\_gt\_emotions\_\_, and no other emotions. Add the sentences at the end of the post. Do not change the words in the post itself. \_\_text\_\_.

The \_\_gt\_emotions\_\_ are replaced with the emotion labels of the GoEmotions dataset, and the \_\_text\_\_ labels are replaced with the input text of the dataset. This prompt consists of three crucial elements. First, we instruct GPT-3.5 and GPT-4 to avoid using emotion labels directly, as this would make the classification task trivial and not generalizable. Second, we promote unambiguity by asking the model to produce text that solely represents the human-annotated emotion, without including any other emotions. Finally, we encourage faithfulness by restricting the context to two sentences. We intentionally do not enforce strict faithfulness, as there are instances in the dataset where the assigned label could be considered outside the range of reasonable interpretations. 

However, this prompting strategy can fail when the emotions are well-suited to the original text. In this case, the context may not improve unambiguity, while failing to be faithful. Hence our context-added input would be worse-aligned to the labels than our original input. In order to prevent this, we introduce an additional type of prompting, called context evaluation, to determine if the original labels are already well-suited for the original text. Our prompt for context evaluation is as follows:

    \textbf{Prompt:} ``system": You are a Reddit user reading posts. ``user" : Is the emotion of \_\_emotion\_\_ well conveyed in this Reddit post? \_\_text\_\_ Answer yes or no.

To enable reproducibility, the entire generation process is done with a temperature of $0$ which means that the generated context is deterministic. In the following sections, we experimentally validate this prompting scheme both with human and model scores.

%% file: dataset.tex
\subsection{Training Dataset Curation}
We add context to the input text in the GoEmotions dataset \cite{goemo}. This dataset consists of Reddit comments which have been annotated for expressed emotions. We choose this particular dataset due to its diverse set of emotion labels. It has 27 emotions along with a neutral label, with sizeable proportions of positive and negative emotions. 

The first set of data we curate is for evaluating the effect of adding context when appropriate. As a baseline, we sample a set of 1000 sentences at random from the training split of the dataset, which we call the Random-Sample Dataset, or \textbf{RS}. Then, we perform context evaluation on the GoEmotions dataset using GPT-X, where X $\in \{3.5, 4\}$. This provides us with two disjoint sets of sentences for each LLM, i.e., sentences with (Context-Present) and without context (Context-Absent). From the Context-Absent sentences, we randomly sub-sample 1000 sentences. We refer to this subset as the GPT-X-Context-Absent Dataset, or \textbf{GPT-X-CA}. The distribution of emotions in GPT-X-CA is shown in Figure~\ref{fig:emo_dist}.
Then, we audit this subset by generating context and appending it to the end of the inputs in GPT-X-CA. We denote this modified dataset as the GPT-X-Context-Absent Modified Dataset, or \textbf{GPT-X-CAM}. We compare RS and GPT-X-CA to evaluate the effectiveness of the context evaluation.

However, when auditing datasets in practice, we also need to consider how to treat the samples which, according to the LLMs, contain enough context. For our first strategy, we add context to every single sentence in RS. This results in the GPT-X-Random-Sample Modified Dataset, or \textbf{GPT-X-RSM}. For our second strategy, we selectively add context only to the Context-Absent sentences in RS to create the GPT-X-Mixed Modified Dataset, or \textbf{GPT-X-MM}. Comparing these datasets will test the hypothesis that adding context to already aligned inputs and labels may hurt empirical performance.


\subsection{External Datasets}
We evaluate the models trained on the subsets defined above on external text-based emotion recognition datasets: \textbf{ISEAR}\cite{isear}, \textbf{SemEval 2019 Task 3}\cite{semeval2019t3}, Emotion Recognition task in \textbf{Tweet Eval}\cite{tweeteval} and \textbf{Daily dialog}\cite{dd}. The ISEAR dataset (International Survey on Emotion Antecedents and Reactions) contains self-reported emotional events and corresponding emotional labels for Joy, Fear, Anger, Sadness, Disgust, Shame, and Guilt. SemEval 2019 Task 3 is composed of sentences from multi-turn dialogues, i.e., each input sentence is composed of the target sentence along with the previous two turns. This dataset is classified into four emotional classes: Happy, Sad, Angry and Others. Emotion Recognition task in Tweet Eval is composed of tweets labelled for four emotion labels: Anger, Joy, Optimism and Sadness. Daily Dialog dataset is composed of multi-turn dialogues between two participants labelled for Neutral, Anger, Disgust, Fear, Happiness, Sadness and Surprise. The label distribution of the datasets is provided in Table~\ref{tab:dataset_eval_stats}.

\input{Tables/emo_rec_dataset}

%% file: Tables/emo_rec_dataset.tex
\begin{table*}[]
\caption{Evaluation Dataset Per Emotion Data Split}
\label{tab:dataset_eval_stats}
\centering
\small
\begin{tabular}{lcccccccccc}
\toprule
                   \multirow{2}{*}{\textbf{Emotions}}& \multicolumn{3}{c}{\textbf{Daily Dialog}}                          & \textbf{ISEAR} & \multicolumn{2}{c}{\textbf{Sem Eval 2019 Task 3}} & \multicolumn{3}{c}{\textbf{Tweet Eval}} & \multirow{2}{*}{\textbf{Mapped GoEmotion}}                  \\ \cmidrule(lr){2-4} \cmidrule(lr){2-4} \cmidrule(lr){5-5} \cmidrule(lr){6-7} \cmidrule(lr){8-10}
                   & \multicolumn{1}{c}{Train} & \multicolumn{1}{c}{Validation} & Test & Total          & \multicolumn{1}{c}{Train}          & Test         & \multicolumn{1}{c}{Train} & \multicolumn{1}{c}{Validation} & Test \\ \midrule
\textbf{Neutral}   & \multicolumn{1}{c}{72143} & \multicolumn{1}{c}{7108}       & 6321 & -              & \multicolumn{1}{c}{-}              & -            & \multicolumn{1}{c}{-}     & \multicolumn{1}{c}{-}          & -    & Neutral  \\ 
\textbf{Anger}     & \multicolumn{1}{c}{827}   & \multicolumn{1}{c}{77}         & 118  & 1069           & \multicolumn{1}{c}{5506}           & 298          & \multicolumn{1}{c}{1400}  & \multicolumn{1}{c}{160}        & 558  & Anger  \\ 
\textbf{Disgust}   & \multicolumn{1}{c}{303}   & \multicolumn{1}{c}{3}          & 47   & 1059           & \multicolumn{1}{c}{-}              & -            & \multicolumn{1}{c}{-}     & \multicolumn{1}{c}{-}          & -    & Disgust  \\ 
\textbf{Fear}      & \multicolumn{1}{c}{146}   & \multicolumn{1}{c}{11}         & 17   & 1096           & \multicolumn{1}{c}{-}              & -            & \multicolumn{1}{c}{-}     & \multicolumn{1}{c}{-}          & -    & Fear  \\ 
\textbf{Happiness} & \multicolumn{1}{c}{11182} & \multicolumn{1}{c}{684}        & 1019 & -              & \multicolumn{1}{c}{4243}           & 284          & \multicolumn{1}{c}{-}     & \multicolumn{1}{c}{-}          & -    &  Joy  \\ 
\textbf{Sadness}   & \multicolumn{1}{c}{969}   & \multicolumn{1}{c}{79}         & 102  & 1074           & \multicolumn{1}{c}{5463}           & 250          & \multicolumn{1}{c}{855}   & \multicolumn{1}{c}{89}         & 382  & Sadness  \\ 
\textbf{Surprise}  & \multicolumn{1}{c}{1600}  & \multicolumn{1}{c}{107}        & 116  & -              & \multicolumn{1}{c}{-}              & -            & \multicolumn{1}{c}{-}     & \multicolumn{1}{c}{-}          & -    & Surprise  \\ 
\textbf{Joy}       & \multicolumn{1}{c}{-}     & \multicolumn{1}{c}{-}          & -    & 1082           & \multicolumn{1}{c}{-}              & -            & \multicolumn{1}{c}{708}   & \multicolumn{1}{c}{97}         & 358  & Joy  \\ 
\textbf{Shame}     & \multicolumn{1}{c}{-}     & \multicolumn{1}{c}{-}          & -    & 1059           & \multicolumn{1}{c}{-}              & -            & \multicolumn{1}{c}{-}     & \multicolumn{1}{c}{-}          & -    &  embarrassment  \\ 
\textbf{Guilt}     & \multicolumn{1}{c}{-}     & \multicolumn{1}{c}{-}          & -    & 1046           & \multicolumn{1}{c}{-}              & -            & \multicolumn{1}{c}{-}     & \multicolumn{1}{c}{-}          & -    &  Remorse  \\ 
\textbf{Optimism}  & \multicolumn{1}{c}{-}     & \multicolumn{1}{c}{-}          & -    & -              & \multicolumn{1}{c}{-}              & -            & \multicolumn{1}{c}{294}   & \multicolumn{1}{c}{28}         & 123  &  Optimism\\ 
\textbf{Others}    & \multicolumn{1}{c}{-}     & \multicolumn{1}{c}{-}          & -    & -              & \multicolumn{1}{c}{14948}          & 4677         & \multicolumn{1}{c}{-}     & \multicolumn{1}{c}{-}          & -    &  All other labels\\ \bottomrule
\end{tabular}
\end{table*}

%% file: experiments.tex
In order to evaluate the effectiveness of adding context to the input text samples, we evaluate with both objective and subjective metrics. In terms of subjective evaluation, we crowdsource responses from Amazon Mechanical Turk (MTurk). We obtain annotations for the degree to which the provided emotional label fits the input sentence. As for the objective evaluation, we show zero-shot performance on a rich set of text-based emotion recognition datasets.

%% file: objectiveEval.tex
We demonstrate the effectiveness of our context evaluation and generation method through a zero-shot, out-of-domain evaluation on the text-based emotion recognition datasets mentioned in Section IV. We use BERT \cite{bert} as the backbone for training the emotion classification models. We decided to freeze the entire BERT model, except the weights in the last encoder layer. Additionally, we add a linear classifier on top of the model. We train identical model instances on each dataset for 20 epochs with a multi-label binary cross-entropy loss. We use the AdamW optimizer and test model performance using F1-macro over 5-fold cross-validation.

\textbf{Context Evaluation}: We demonstrate the effectiveness of our context evaluation prompting by comparing the F1-macro scores of GPT-X-CA and RS datasets. We would expect the GPT-X-CA performance to be lower since these sentences were determined to not contain context according to our context evaluation prompting.

\textbf{Context Generation}: The emotion labels for the utilized evaluation datasets (see Section IV) do not always intersect with those of GoEmotions. Hence, we map the GoEmotion labels to the respective evaluation dataset, as shown in Table~\ref{tab:dataset_eval_stats}. Additionally, we evaluate sentiment analysis using Stanford Sentiment Treebank~\cite{sst} and SemEval 2017 Task 4\cite{semeval2017t4} using the zero-shot paradigm by mapping emotions to positive, negative and neutral.

%% file: subjectiveEval.tex
We use subjective evaluation in order to verify whether the added context helps to express the labeled emotion better. We perform the evaluation on the GPT-4-CA and GPT-4-CAM datasets. To test for statistical significance, we choose a subset of emotions from the Go Emotions dataset, i.e., the 4 most common positive and negative emotions along with the neutral emotion for a total of 9 emotion labels. The resulting set of emotions contains admiration, love, approval, amusement, neutral, annoyance, anger, sadness and disapproval.

We perform an MTurk survey where each participant is given a set of 20 questions consisting of a random split of GPT-4-CA and GPT-4-CAM. The only criteria for participants was being proficient in English. The survey is between subjects where no single participant is presented with the GPT4-CA and GPT4-CAM version of the same prompt. For each question, the participants choose on a 5-point Likert scale the extent to which the labeled emotion is expressed in the text. We do not expect our data to follow a normal distribution, hence we perform a Kruksal-Wallis test \cite{kwt} and a post-hoc Dunn test \cite{dunntest} to determine the statistical significance between the set of chosen emotions. For the Dunn test, we use the Benjamini-Hochberg adjustment \cite{benjamini}.

\begin{table}[]
\caption{Statistical descriptors per emotion for GPT4-CA and GPT4-CAM Subjective Evaluation, text in bold indicates higher mean}
\label{tab:emotion_wise_mean_std}
\centering
\small
\begin{tabular}{@{}lcccccc@{}}
\toprule
\multirow{2}{*}{\textbf{Emotions}} & \multicolumn{2}{c}{\textbf{Mean}} & \multicolumn{2}{c}{\textbf{Std Dev.}} & \multicolumn{2}{c}{\textbf{Median}} \\ \cmidrule(lr){2-3} \cmidrule(lr){4-5} \cmidrule(lr){6-7}
& CA & CAM & CA & CAM & CA & CAM \\ \midrule
\textbf{admiration} & 2.93 & \textbf{3.71} & 0.89 & 0.73 & 3 & 4 \\
\textbf{love} & 2.93 & \textbf{3.71} & 1.01 & 0.89 & 3 & 4 \\
\textbf{approval} & 2.74 & \textbf{3.64} & 1.09 & 0.72 & 3 & 4 \\
\textbf{neutral} & 2.68 & \textbf{3.13} & 0.81 & 0.86 & 3 & 3 \\
\textbf{amusement} & 3.10 & \textbf{3.46} & 0.90 & 0.73 & 3 & 3 \\
\textbf{annoyance} & 3.06 & \textbf{3.48} & 0.78 & 0.97 & 3 & 4 \\
\textbf{anger} & 3.01 & \textbf{3.67} & 0.94 & 0.87 & 3 & 4 \\
\textbf{sadness} & 2.95 & \textbf{3.59} & 0.69 & 0.85 & 3 & 4 \\
\textbf{disapproval} & 2.44 & \textbf{3.00} & 0.74 & 0.75 & 3 & 4 \\ \bottomrule
\end{tabular}
\end{table}

%% file: Tables/results_bert_sent.tex
\begin{table}[]
\caption{Sentiment Analysis Evaluation, text in bold indicates higher performance}
\label{tab:sent_results}
\centering
\begin{tabular}{cccccc}
\toprule
\textbf{Dataset}    & \multicolumn{2}{c}{\textbf{SST}}                        & \multicolumn{3}{c}{\textbf{Sem Eval 2017}}                                                    \\ \cmidrule(lr){1-1} \cmidrule(lr){2-3} \cmidrule(lr){4-6}
\textbf{Model}      & \multicolumn{1}{c}{\textbf{Train}} & \textbf{Validation} & \multicolumn{1}{c}{\textbf{Train}} & \multicolumn{1}{c}{\textbf{Validation}} & \textbf{Test} \\ \midrule
\textbf{GPT-4-CA}   & \multicolumn{1}{c}{65.2}           & 61.2                & \multicolumn{1}{c}{\textbf{55.1}}  & \multicolumn{1}{c}{\textbf{55.7}}       & \textbf{52.2} \\ 
\textbf{GPT-4-CAM}   & \multicolumn{1}{c}{\textbf{68.3}}  & \textbf{67.5}       & \multicolumn{1}{c}{48.7}           & \multicolumn{1}{c}{47.4}                & 50.3          \\ \midrule
\textbf{GPT-3.5-CA} & \multicolumn{1}{c}{64.3}           & 61.1                & \multicolumn{1}{c}{\textbf{54.1}}  & \multicolumn{1}{c}{\textbf{54.4}}       & 49.6          \\ 
\textbf{GPT-3.5-CAM} & \multicolumn{1}{c}{\textbf{66.7}}  & \textbf{64.9}       & \multicolumn{1}{c}{49.1}           & \multicolumn{1}{c}{48.2}                & \textbf{50.3} \\ \bottomrule
\end{tabular}
\end{table}

%% file: results.tex
\subsection{Objective Evaluation}
\subsubsection{Context Evaluation}
From Figure.\ref{fig:cont_eval}, we can note that the performance of the GPT-X-CA datasets is consistently lower compared to RS dataset. The difference between the datasets is that the CA dataset only has samples which the language model determined was lacking context, and the RS dataset contains a random mix of samples that lack and do not lack context. Thus, the lower performance is indicative of the context evaluation capabilities of GPT-3.5 and GPT-4.

\input{Tables/context_examples} 

\subsubsection{Context Generation}
Table~\ref{tab:context_examples} shows some examples of context generated by GPT-4 along with the corresponding GoEmotions input and its label. 

\begin{figure*}[htbp]
  \centering
  \subfigure[Context Evaluation Validation]{
    \includegraphics[width=\linewidth]{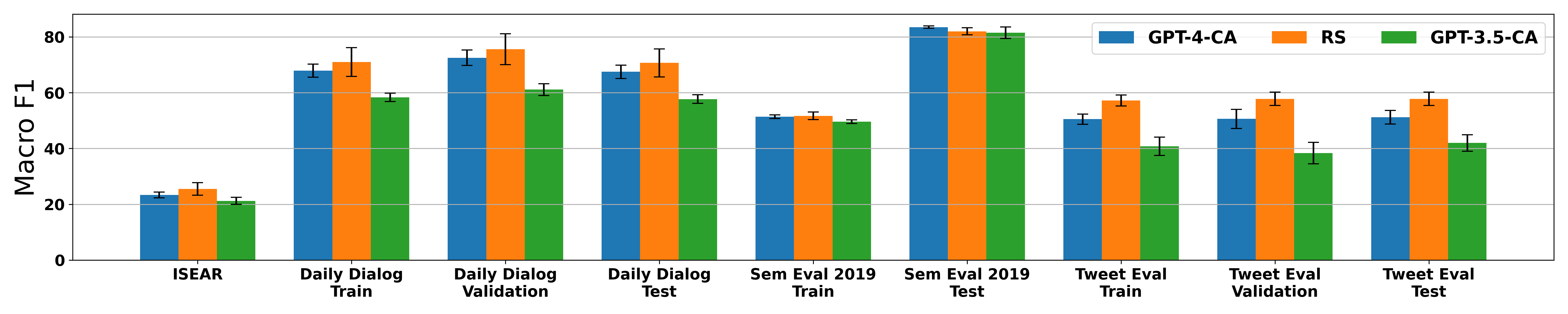}
    \label{fig:cont_eval}
  }
  \subfigure[GPT-3.5 Macro F1 Scores]{
    \includegraphics[width=\linewidth]{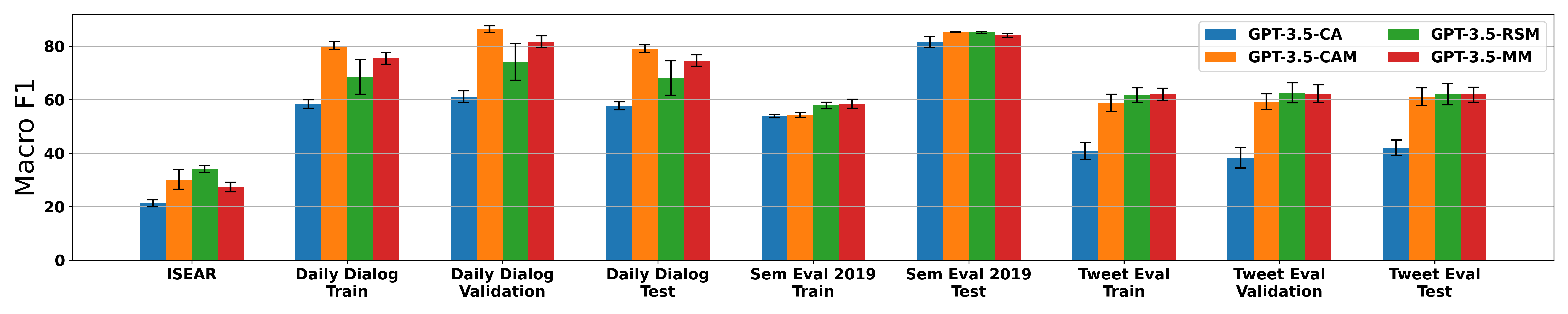}
    \label{fig:gpt3_cont_eval}
  }
  \subfigure[GPT-4 Macro F1 Scores]{
    \includegraphics[width=\linewidth]{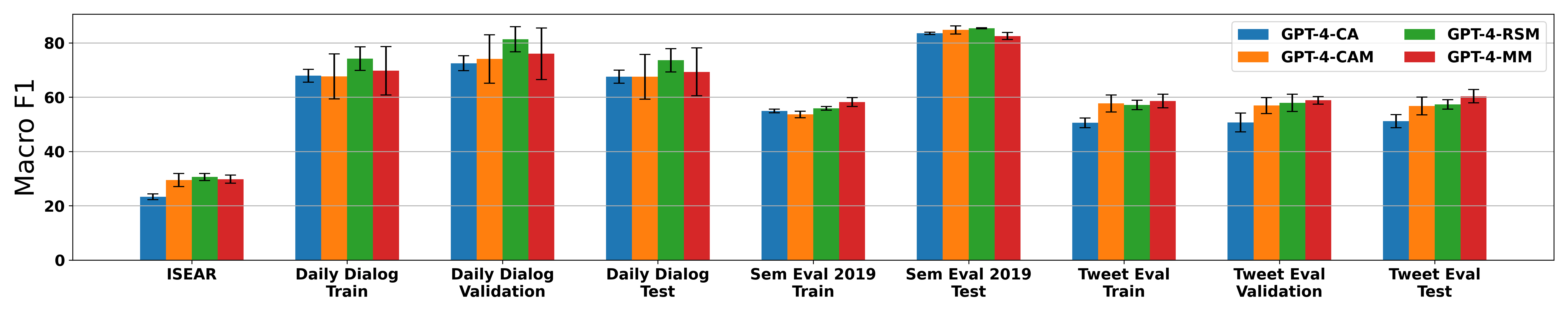}
    \label{fig:gpt4_cont_eval}
  }
  \caption{Macro F1 scores and standard deviations across folds for evaluation on text-based emotion recognition datasets}
  \label{fig:whole_figure}
\end{figure*}

From Figure.\ref{fig:gpt3_cont_eval} and Figure.\ref{fig:gpt4_cont_eval}, we see a noticeable improvement in the performance of the model trained on the GPT-X-CAM dataset compared to GPT-X-CA for most tasks. This validates the hypothesis that adding relevant context to input sentences leads to better generalization capabilities in downstream tasks such as emotion recognition. Despite evaluating on substantially different domains such as in the case of Tweet Eval, which has emotion labels for tweets or in the case of the Daily Dialog dataset wherein the input text is from multi-turn dialog akin to everyday conversations, we observe that context-enhanced models perform better compared to the models which are trained on inputs without enhanced context.

With regard to the ISEAR dataset, we identify that the performance of all our models is subpar. This could be attributed to a couple of factors. Firstly, this dataset is composed of intended emotions and corresponding textual descriptions of the experiences from the authors themselves. At the same time, the training dataset (GoEmotions) is annotated by a third-party. We note that third-party and first-party respondents attend to distinct constructs while labelling emotions\cite{lens} which might make it difficult for models trained on third-party labeled datasets to generalize to first-party labeled datasets. Also, ISEAR has labels such as shame and guilt which are mapped to embarrassment and remorse, respectively. There are few samples with these labels in the curated GoEmotions subset, making it harder for the model to classify them.

Additionally, GPT-X-RSM and GPT-X-MM models outperform RS in most cases. This validates our hypothesis that context-enhanced prompting is effective for improving the alignment between inputs and emotional labels. We do not see a distinct trend between the performance of the GPT-X-RM and GPT-X-MM models. This means that there is no clear advantage or disadvantage of enhancing both Context-Present and Context-Absent sentences, versus selectively enhancing Context-Absent sentences.

As OpenAI has yet to fully provide architecture and training details on the difference between GPT-3.5 and GPT-4, we do not compare across LLMs.

\subsubsection{Sentiment Analysis}
For sentiment analysis,  we do not observe such consistent performance gains as seen in Table~\ref{tab:sent_results}. Firstly, emotion recognition is a more challenging task compared to sentiment analysis. Hence, sentiment analysis may not require as fine-grained context as that of emotion recognition. Therefore, the enchanced context helps disambiguate the granularity of various emotions. However, it does not offer other additional benefit when the boundaries are more distinct such as in positive, negative, and neutral in terms of sentiment. 

\subsection{Subjective Evaluation}

The survey results show a higher mean for the GPT4-CAM dataset compared to the GPT4-CA accross all emotions. The average, standard deviation and median of all responses are reported in Table~\ref{tab:emotion_wise_mean_std}. We perform Kruksal-Wallis test on the responses, and the parameters chosen are alpha ($\alpha=0.05$) and power ($1-\beta=0.95$). Given a sample size of $N = 1977$, we obtain \textit{\textbf{p\textless0.001}} and moderate effect size ($\eta^2=0.08$). The Kruksal-Wallis test was followed by the pairwise post-hoc Dunn test with Benjamini-Hochberg adjustment to determine statistical significance per emotion. As stated previously, we consider a set of 9 emotions, which would yield 18 groups, i.e., for each emotion, we have one group for GPT-4-CA and another for GPT-4-CAM. We obtain statistical significance between GPT-4-CA and GPT-4-CAM for admiration (\textit{\textbf{p\textless{}0.001}}), love (\textit{\textbf{p\textless{}0.001}}), approval (\textit{\textbf{p\textless{}0.001}}), neutral (\textit{\textbf{p=0.0272}}), annoyance (\textit{\textbf{$p=0.024$}}), sadness (\textit{\textbf{p=0.015}}) and anger (\textit{\textbf{p=0.005}}). For these emotions, the results show that from a human perspective, our added context better aligns the input text to their human-annotated labels. We do not observe a statistical significance for amusement (\textit{p=0.052}) and disapproval (\textit{p=0.191}).

\subsection{Context Analysis}
We compare the distribution of words for the generated context in GPT-4-CAM and the original GoEmotions text in GPT-4-CA. For admiration, the top 3 words in the original text are ``good" with frequency $4.6\%$, ``great" with frequency $3.4\%$, and ``one" with frequency $2.0\%$. The top 3 words in the generated context are ``truly" with frequency $14.2\%$, ``say" with frequency $7.4\%$, and ``awe" with the frequency of $5.6\%$. For the label of anger, the top 3 words in the original text are ``hate" with a frequency of $3.5\%$, ``just" with the frequency of $2.4\%$, and an expletive with frequency ``1.8\%". The top 3 words in the generated context are ``infuriating" with frequency $8.4\%$, ``believe" with frequency $5.1\%$,  and ``just" with frequency $5.1\%$. For both emotions, we note that the generated context contains words with frequencies that are much higher than that of GoEmotions. This means that LLMs fail to generate context with the same diversity as in-the-wild text. Although our empirical results do not favor either GPT-4-RSM or GPT-4-MM over the other, our analysis shows that GPT-4-MM may be the better prompting strategy because its sample distribution is closer to that of in-the-wild text data.

%% file: Tables/context_examples.tex

\begin{table*}[]
\caption{Examples of GPT-4 context generation}
\label{tab:context_examples}
\centering
\begin{tabular}{ccc}
\toprule
\textbf{Text Input}                                                                                                                                                                  & \textbf{Context}                                                                                                                                                                                                                              & \textbf{Label} \\ \midrule
What do the {[}NAME{]}  have to do with it?                                                                                                                                          & \begin{tabular}[c]{@{}c@{}}I can't help but wonder how they're connected to this situation. \\ Can anyone shed some light on this intriguing aspect?\end{tabular}                                                                             & curiosity      \\ \hline
Maaaan, that woman is crazy!  And the straightjacket type!                                                                                                                           & \begin{tabular}[c]{@{}c@{}}However, it's impossible not to respect her unwavering dedication and passion \\ for what she does. She truly stands out as a remarkable individual in her field.\end{tabular}                                     & admiration     \\ \hline
\begin{tabular}[c]{@{}c@{}}Someone who seems great while texting might not \\ be that great in person. You can't make someone \\ meet a stranger if they don't want to.\end{tabular} & \begin{tabular}[c]{@{}c@{}}I must say, it's quite concerning how some people can be so deceptive \\ in their online interactions. It's important to be cautious and not blindly trust \\ everyone we come across on the internet\end{tabular} & disapproval    \\ \bottomrule
\end{tabular}
\end{table*}

%% file: conclusion.tex
In this paper, we introduce a framework for defining emotional context for text and provide an algorithm to identify sentences that lack this context. We also demonstrate that prompting LLMs to enrich these sentences with context results in significantly improved performance for out-of-domain datasets. Additionally, these enriched sentences show an increased alignment with labelled emotions in subjective evaluation, validating our prompting approach and applicability of this technique to real world datasets. 

For future work, we believe our dataset augmentation strategy can be explored at a greater scale. We hypothesize that by fully auditing GoEmotions and other datasets, our method can be used to improve the state-of-the art in text-based emotion classification. In addition to increasing the number of samples in the audited dataset that we consider, we also believe that prompting can be used as a data augmentation strategy with a higher temperature setting.

Additionally, we believe our prompting strategy can be extended. We believe that more fine-grained knowledge of the target audience, such as the subreddit of a particular input text, can potentially qualify our definition of context and make our prompts more effective and diverse. Furthermore, we believe that our prompting strategy can be expanded beyond text-based emotion classification to more modalities, more languages, and other tasks with label ambiguity.

%% file: ethicalImpact.tex
We note that LLMs do not have abstract constructs of emotions, but merely have pattern-based understanding of emotional words from large scale corpora in training. Therefore, we can not accurately infer the intended first party emotions from text but merely model the emotion expressed based on semantics in the text. Hence, there will be cases wherein the intended emotions differ from the emotions which are expressed in the text.

Furthermore, the text generated from LLMs mirror biases in their training data. This means our auditing methods may introduce such biases into text-based emotion classification datasets, and the models trained on top of them. Currently, our work does not explore the implications of the biases that our generated datasets might have.

%% file: ACII2023-template.bbl
\begin{thebibliography}{10}
\providecommand{\url}[1]{#1}
\csname url@samestyle\endcsname
\providecommand{\newblock}{\relax}
\providecommand{\bibinfo}[2]{#2}
\providecommand{\BIBentrySTDinterwordspacing}{\spaceskip=0pt\relax}
\providecommand{\BIBentryALTinterwordstretchfactor}{4}
\providecommand{\BIBentryALTinterwordspacing}{\spaceskip=\fontdimen2\font plus
\BIBentryALTinterwordstretchfactor\fontdimen3\font minus
  \fontdimen4\font\relax}
\providecommand{\BIBforeignlanguage}[2]{{%
\expandafter\ifx\csname l@#1\endcsname\relax
\typeout{** WARNING: IEEEtran.bst: No hyphenation pattern has been}%
\typeout{** loaded for the language `#1'. Using the pattern for}%
\typeout{** the default language instead.}%
\else
\language=\csname l@#1\endcsname
\fi
#2}}
\providecommand{\BIBdecl}{\relax}
\BIBdecl

\bibitem{isear}
K.~R. Scherer and H.~G. Wallbott, ``Evidence for universality and cultural
  variation of differential emotion response patterning.'' \emph{Journal of
  personality and social psychology}, vol.~66, no.~2, p. 310, 1994.

\bibitem{dd}
Y.~Li, H.~Su, X.~Shen, W.~Li, Z.~Cao, and S.~Niu, ``Dailydialog: A manually
  labelled multi-turn dialogue dataset,'' in \emph{Proceedings of The 8th
  International Joint Conference on Natural Language Processing (IJCNLP 2017)},
  2017.

\bibitem{tweeteval}
\BIBentryALTinterwordspacing
F.~Barbieri, J.~Camacho-Collados, L.~Espinosa~Anke, and L.~Neves,
  ``{T}weet{E}val: Unified benchmark and comparative evaluation for tweet
  classification,'' in \emph{Findings of the Association for Computational
  Linguistics: EMNLP 2020}.\hskip 1em plus 0.5em minus 0.4em\relax Online:
  Association for Computational Linguistics, Nov. 2020, pp. 1644--1650.
  [Online]. Available: \url{https://aclanthology.org/2020.findings-emnlp.148}
\BIBentrySTDinterwordspacing

\bibitem{semeval2019t3}
A.~Chatterjee, K.~N. Narahari, M.~Joshi, and P.~Agrawal, ``{S}em{E}val-2019
  task 3: {E}mo{C}ontext contextual emotion detection in text,'' in
  \emph{Proceedings of the 13th International Workshop on Semantic
  Evaluation}.\hskip 1em plus 0.5em minus 0.4em\relax Minneapolis, Minnesota,
  USA: Association for Computational Linguistics, Jun. 2019, pp. 39--48.

\bibitem{goemo}
D.~Demszky, D.~Movshovitz-Attias, J.~Ko, A.~Cowen, G.~Nemade, and S.~Ravi,
  ``{GoEmotions: A Dataset of Fine-Grained Emotions},'' in \emph{58th Annual
  Meeting of the Association for Computational Linguistics (ACL)}, 2020.

\bibitem{gpt4}
OpenAI, ``Gpt-4 technical report,'' 2023.

\bibitem{10.5555/3295222.3295349}
A.~Vaswani, N.~Shazeer, N.~Parmar, J.~Uszkoreit, L.~Jones, A.~N. Gomez,
  L.~Kaiser, and I.~Polosukhin, ``Attention is all you need,'' in
  \emph{Proceedings of the 31st International Conference on Neural Information
  Processing Systems}, ser. NIPS'17.\hskip 1em plus 0.5em minus 0.4em\relax Red
  Hook, NY, USA: Curran Associates Inc., 2017, p. 6000–6010.

\bibitem{gpt3}
T.~B. Brown, B.~Mann, N.~Ryder, M.~Subbiah, J.~Kaplan, P.~Dhariwal,
  A.~Neelakantan, P.~Shyam, G.~Sastry, A.~Askell, S.~Agarwal, A.~Herbert-Voss,
  G.~Krueger, T.~Henighan, R.~Child, A.~Ramesh, D.~M. Ziegler, J.~Wu,
  C.~Winter, C.~Hesse, M.~Chen, E.~Sigler, M.~Litwin, S.~Gray, B.~Chess,
  J.~Clark, C.~Berner, S.~McCandlish, A.~Radford, I.~Sutskever, and D.~Amodei,
  ``Language models are few-shot learners,'' 2020.

\bibitem{instructgpt}
L.~Ouyang, J.~Wu, X.~Jiang, D.~Almeida, C.~L. Wainwright, P.~Mishkin, C.~Zhang,
  S.~Agarwal, K.~Slama, A.~Ray, J.~S.~J. Hilton, F.~Kelton, L.~Miller,
  M.~Simens, A.~Askell, P.~Welinder, P.~Christiano, and R.~L. Jan~Leike,
  ``Aligning language models to follow instructions,'' 2022.

\bibitem{bert}
\BIBentryALTinterwordspacing
J.~Devlin, M.-W. Chang, K.~Lee, and K.~Toutanova, ``{BERT}: Pre-training of
  deep bidirectional transformers for language understanding,'' in
  \emph{Proceedings of the 2019 Conference of the North {A}merican Chapter of
  the Association for Computational Linguistics: Human Language Technologies,
  Volume 1 (Long and Short Papers)}.\hskip 1em plus 0.5em minus 0.4em\relax
  Minneapolis, Minnesota: Association for Computational Linguistics, Jun. 2019,
  pp. 4171--4186. [Online]. Available: \url{https://aclanthology.org/N19-1423}
\BIBentrySTDinterwordspacing

\bibitem{jackofalltradeschatgpt}
J.~Kocoń, I.~Cichecki, O.~Kaszyca, M.~Kochanek, D.~Szydło, J.~Baran,
  J.~Bielaniewicz, M.~Gruza, A.~Janz, K.~Kanclerz, A.~Kocoń, B.~Koptyra,
  W.~Mieleszczenko-Kowszewicz, P.~Miłkowski, M.~Oleksy, M.~Piasecki, Łukasz
  Radliński, K.~Wojtasik, S.~Woźniak, and P.~Kazienko, ``Chatgpt: Jack of all
  trades, master of none,'' 2023.

\bibitem{emotionx}
Y.-H. Huang, S.-R. Lee, M.-Y. Ma, Y.-H. Chen, Y.-W. Yu, and Y.-S. Chen,
  ``Emotionx-idea: Emotion bert – an affectional model for conversation,''
  2022.

\bibitem{memobert}
J.~Zhao, R.~Li, Q.~Jin, X.~Wang, and H.~Li, ``Memobert: Pre-training model with
  prompt-based learning for multimodal emotion recognition,'' 2021.

\bibitem{TER_Challenges}
J.~Deng and F.~Ren, ``A survey of textual emotion recognition and its
  challenges,'' \emph{IEEE Transactions on Affective Computing}, vol.~14,
  no.~1, pp. 49--67, 2023.

\bibitem{gilardi2023chatgpt}
F.~Gilardi, M.~Alizadeh, and M.~Kubli, ``Chatgpt outperforms crowd-workers for
  text-annotation tasks,'' 2023.

\bibitem{barrett2019emotional}
L.~F. Barrett, R.~Adolphs, S.~Marsella, A.~M. Martinez, and S.~D. Pollak,
  ``Emotional expressions reconsidered: Challenges to inferring emotion from
  human facial movements,'' \emph{Psychological science in the public
  interest}, vol.~20, no.~1, pp. 1--68, 2019.

\bibitem{mittal2020emoticon}
T.~Mittal, P.~Guhan, U.~Bhattacharya, R.~Chandra, A.~Bera, and D.~Manocha,
  ``Emoticon: Context-aware multimodal emotion recognition using frege’s
  principle, in 2020 ieee,'' in \emph{CVF Conference on Computer Vision and
  Pattern Recognition (CVPR) pp}, 2020, pp. 14\,222--14\,231.

\bibitem{lee2019context}
J.~Lee, S.~Kim, S.~Kim, J.~Park, and K.~Sohn, ``Context-aware emotion
  recognition networks,'' in \emph{Proceedings of the IEEE/CVF international
  conference on computer vision}, 2019, pp. 10\,143--10\,152.

\bibitem{kosti2017emotion}
R.~Kosti, J.~M. Alvarez, A.~Recasens, and A.~Lapedriza, ``Emotion recognition
  in context,'' in \emph{Proceedings of the IEEE conference on computer vision
  and pattern recognition}, 2017, pp. 1667--1675.

\bibitem{liscombe2005using}
J.~Liscombe, G.~Riccardi, and D.~Hakkani-Tur, ``Using context to improve
  emotion detection in spoken dialog systems,'' 2005.

\bibitem{Lee2005Towarddetectingemotionsin}
C.~M. Lee and S.~S. Narayanan, ``Toward detecting emotions in spoken dialogs,''
  \emph{IEEE Transactions on Speech and Audio Processing}, vol.~13, no.~2, pp.
  293--303, mar 2005.

\bibitem{zhang2019modeling}
D.~Zhang, L.~Wu, C.~Sun, S.~Li, Q.~Zhu, and G.~Zhou, ``Modeling both
  context-and speaker-sensitive dependence for emotion detection in
  multi-speaker conversations.'' in \emph{IJCAI}, 2019, pp. 5415--5421.

\bibitem{confident}
C.~Northcutt, L.~Jiang, and I.~Chuang, ``Confident learning: Estimating
  uncertainty in dataset labels,'' \emph{J. Artif. Int. Res.}, vol.~70, p.
  1373–1411, may 2021.

\bibitem{laion5b}
C.~Schuhmann, R.~Beaumont, R.~Vencu, C.~W. Gordon, R.~Wightman, M.~Cherti,
  T.~Coombes, A.~Katta, C.~Mullis, M.~Wortsman, P.~Schramowski, S.~R.
  Kundurthy, K.~Crowson, L.~Schmidt, R.~Kaczmarczyk, and J.~Jitsev,
  ``{LAION}-5b: An open large-scale dataset for training next generation
  image-text models,'' in \emph{Thirty-sixth Conference on Neural Information
  Processing Systems Datasets and Benchmarks Track}, 2022.

\bibitem{clip}
A.~Radford, J.~W. Kim, C.~Hallacy, A.~Ramesh, G.~Goh, S.~Agarwal, G.~Sastry,
  A.~Askell, P.~Mishkin, J.~Clark, G.~Krueger, and I.~Sutskever, ``Learning
  transferable visual models from natural language supervision,'' \emph{CoRR},
  vol. abs/2103.00020, 2021.

\bibitem{sst}
R.~Socher, A.~Perelygin, J.~Wu, J.~Chuang, C.~D. Manning, A.~Ng, and C.~Potts,
  ``Recursive deep models for semantic compositionality over a sentiment
  treebank,'' in \emph{Proceedings of the 2013 Conference on Empirical Methods
  in Natural Language Processing}.\hskip 1em plus 0.5em minus 0.4em\relax
  Seattle, Washington, USA: Association for Computational Linguistics, Oct.
  2013, pp. 1631--1642.

\bibitem{semeval2017t4}
S.~Rosenthal, N.~Farra, and P.~Nakov, ``Semeval-2017 task 4: Sentiment analysis
  in twitter,'' in \emph{Proceedings of the 11th international workshop on
  semantic evaluation (SemEval-2017)}, 2017, pp. 502--518.

\bibitem{kwt}
W.~H. Kruskal and W.~A. Wallis, ``Use of ranks in one-criterion variance
  analysis,'' \emph{Journal of the American statistical Association}, vol.~47,
  no. 260, pp. 583--621, 1952.

\bibitem{dunntest}
O.~J. Dunn, ``Multiple comparisons using rank sums,'' \emph{Technometrics},
  vol.~6, no.~3, pp. 241--252, 1964.

\bibitem{benjamini}
W.~Haynes, ``Benjamini–{Hochberg} {Method},'' in \emph{Encyclopedia of
  {Systems} {Biology}}, W.~Dubitzky, O.~Wolkenhauer, K.-H. Cho, and H.~Yokota,
  Eds.\hskip 1em plus 0.5em minus 0.4em\relax New York, NY: Springer New York,
  2013, pp. 78--78.

\bibitem{lens}
R.~Gifford, ``A lens-mapping framework for understanding the encoding and
  decoding of interpersonal dispositions in nonverbal behavior.'' \emph{Journal
  of Personality and Social Psychology}, vol.~66, pp. 398--412, 1994, place: US
  Publisher: American Psychological Association.

\end{thebibliography}
